\pdfoutput=1

\documentclass[11pt]{article}

\usepackage[final]{acl}

\usepackage{times}
\usepackage{latexsym}

\usepackage[T1]{fontenc}

\usepackage[utf8]{inputenc}

\usepackage{microtype}

\usepackage{inconsolata}

\usepackage{graphicx}
\usepackage{subcaption}
\usepackage{booktabs}
\usepackage{multirow}
\usepackage{amsmath}

\usepackage[most]{tcolorbox}

\title{
REInstruct: Building Instruction Data from Unlabeled Corpus
}

\author{
  Shu Chen${}^{1,3}$,
  Xinyan Guan${}^{1,3}$,
  Yaojie Lu${}^{1,}$\thanks{~ Corresponding authors.},
  Hongyu Lin${}^{1}$,
  Xianpei Han${}^{1,2,4,}$\footnotemark[1],
  Le Sun${}^{1,2,4}$
  \\
  ${}^{1}$Chinese Information Processing Laboratory ~
  ${}^{2}$State Key Laboratory of Computer Science \\
  Institute of Software, Chinese Academy of Sciences\\
  ${}^{3}$University of Chinese Academy of Sciences \\
  ${}^{4}$Key Laboratory of System Software, Chinese Academy of Sciences \\
 {\tt \{chenshu2020,guanxinyan2022,luyaojie,hongyu,xianpei,sunle\}@iscas.ac.cn}
}

\begin{document}
\maketitle
\begin{abstract}

Manually annotating instruction data for large language models is 
difficult, costly, and hard to scale.
Meanwhile, current automatic annotation methods typically rely on distilling synthetic data from proprietary LLMs, which not only limits the upper bound of the quality of the instruction data but also raises potential copyright issues.
In this paper, we propose REInstruct, a simple and scalable method to automatically build instruction data from an unlabeled corpus without heavy reliance on proprietary LLMs and human annotation.
Specifically, REInstruct first selects a subset of unlabeled texts that potentially contain well-structured helpful and insightful content and then generates instructions for these texts.
To generate accurate and relevant responses for effective and robust training, REInstruct further proposes a rewriting-based approach to improve the quality of the generated instruction data.
By training Llama-7b on a combination of 3k seed data and 32k synthetic data from REInstruct, fine-tuned model achieves a 65.41\% win rate on AlpacaEval leaderboard against text-davinci-003, outperforming other open-source, non-distilled instruction data construction methods.
The code is publicly available at \url{https://github.com/cs32963/REInstruct}.
\end{abstract}

\section{Introduction}

Instruction Tuning \cite{instruct-gpt} is an important step in aligning large language models (LLMs) \cite{gpt3,llama}. By tuning LLMs on <instruction, response> pairs, instruction tuning enhances zero-shot performance of LLMs across a variety of tasks \cite{flan,flan-t5}, and paves the way for further preference training such as reinforcement learning from human feedback (RLHF) \cite{instruct-gpt,llama2} and direct preference optimization (DPO) \cite{dpo}.

However, it is very difficult and costly to manually annotate high-quality instruction data \cite{oasst}. Previous methods for automatic annotation largely rely on powerful proprietary LLMs like ChatGPT \cite{chatgpt}, using techniques such as Self-Instruct \cite{self-instruct} and Evol-Instruct \cite{wizardlm}. While these methods can generate large amount of instruction data and facilitate research in instruction tuning \cite{what-makes-good-data-for-alignment,alpagasus}, the upper bound of the quality of generated data is limited by the power of proprietary LLMs.

\begin{figure}
  \centering
  \includegraphics[width=0.9\columnwidth]{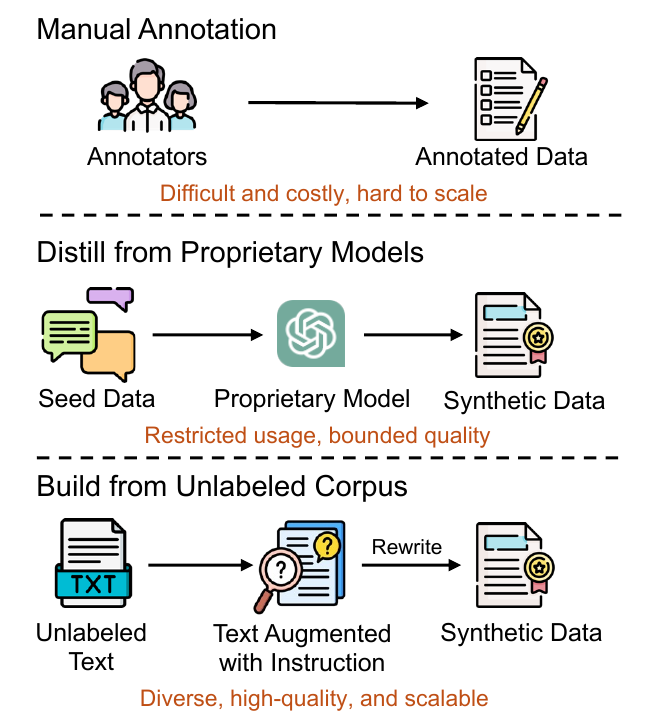}
  \caption{Comparison of our automatic instruction data annotation method and previous methods.}
  \label{fig:intro}
\end{figure}

In this paper, we propose a simple and scalable method to automatically build instruction data from unlabeled text corpus. The proposed method requires only small amount of human-labeled data, and does not rely heavily on powerful proprietary LLMs. Our key motivation is that unlabeled text corpus contains a subset which is filled with high-quality instruction following information, and converting these texts into proper instruction following format is much easier than labeling them from scratch. Specifically, we first employ a set of heuristic rules to obtain unlabeled text containing well-structured helpful and insightful content. Given these content, we then generate related instruction to these high-quality content using an LLM reversely trained on a small set of high-quality human-annotated seed instruction data. Finally, given the generated instruction and the content,  we employ a response rewriting process on the content, in order to obtain the high-quality and accurate response of the instruction from the content. To this end, we leverage a rewriting process in the form of generative machine reading comprehension, which is conducted by an addtional LLM trained on a small set of rewriting data.

To demonstrate the effectiveness of our automatic labeling method, we train LLMs jointly on our synthetic data and seed data across different data scales. We find that, Llama-7b \cite{llama} model finetuned with our largest 32k synthetic data and 3k seed data achieves 65.41\% win rate on AlpacaEval leaderboard, outperforming other open-source, non-distilled methods with the same pre-trained checkpoint. This demonstrate the effectiveness of REInstruct.

Generally, the main contributions of this paper are:
\begin{itemize}
    \item We propose a method to automatically build high-quality instruction data from unlabeled corpus, which requires only a small set of seed instruction data and a small set of rewriting data. 
    \item We identify several heuristic rules to obtain high-quality unlabeled text for constructing instruction-following data, and propose a generative machine reading comprehension-based rewriting process to obtain high-quality response for each instruction.
\end{itemize}

\begin{figure*}[ht]
  \centering
  \includegraphics[width=\textwidth]{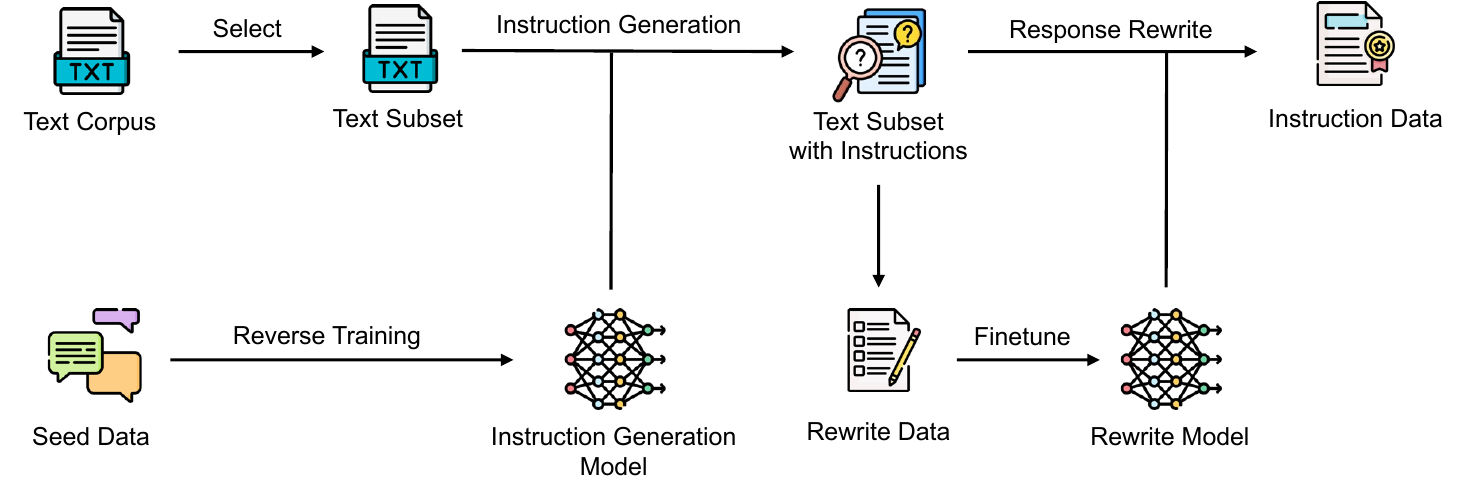}
  \caption{The overall framework of our instruction data annotation method.}
  \label{fig:method}
\end{figure*}

\section{Related Work}

\textbf{Instruction Tuning.} Early works on instruction tuning focus on traditional NLP tasks, and instruction tuned LLMs show improved zero-shot performance across a variety of NLP tasks \cite{flan, flan-t5, natural-instructions, t0}. Since the release of ChatGPT \cite{chatgpt}, there have been increasing interests in general purpose instruction tuning, which expands instructions beyond NLP tasks \cite{instruct-gpt,self-instruct, alpaca, vicuna}. General purpose instruction tuning serves as the first step to build powerful LLM-based AI assistant \cite{instruct-gpt,chatgpt,gpt4}, providing foundation for further alignment training such as RLHF \cite{instruct-gpt} and DPO \cite{dpo}.

\textbf{Automatic Instruction Data labeling.} Manually annotating instruction data is very difficult and expensive, making it hard to scale \cite{oasst}. Early instruction datasets are obtained by converting existing task-specific NLP datasets into instruction format \cite{flan,flan-t5,t0}. Recent works have largely rely on powerful proprietary LLMs \cite{instruct-gpt,chatgpt} to generate high-quality instruction data. Alpaca \cite{alpaca} use Self-Instruct \cite{self-instruct} to collect instruction data from InstructGPT \cite{instruct-gpt} using its in-context learning ability \cite{gpt3}. WizardLM \cite{wizardlm} devise Evol-Instruct to generate more complex and difficult instructions by prompting ChatGPT \cite{chatgpt}. While these works \cite{ultrachat,vicuna,lamini} can generate large amount of instruction data and help facilitate research in instruction tuning \cite{what-makes-good-data-for-alignment,alpagasus}, they rely on the underlying proprietary LLMs to ensure the diversity, difficulty and quality of instruction data, and fail to address the difficulty of annotating instruction data for these powerful proprietary LLMs.

There have been some effort in automatic instruction dataset annotation without reliance on powerful proprietary LLMs, trying to build instruction data using unlabeled text \cite{udit}. Closest to our work is Humpback \cite{self-alignment}, which proposes to add instructions to unlabeled text and employ an iterative self-rating and selection pipeline to curate high-quality instruction data. However, Humpback relies on the existence of high quality unlabeled text already in the form of instruction response, and leaves the noise in unlabeled text and during instruction generation unaddressed, whereas we tackle the noise directly, resulting in a more efficient and robust pipeline.

\textbf{Simulated Human Supervision.} Since human supervision is difficult and expensive to obtain, recent works have explored the possibility of training LLMs to generate pseudo human supervisions, which usually helps reduce the annotation cost and speed up the alignment process. RLHF \cite{instruct-gpt,llama2} train LLMs on human preference data to provide reward signals for reinforcement learning to align LLMs. CoachLM \cite{coachlm} use expert-annotated instruction revision data to train an LLM for automatic instruction data optimization. Our work is related as we train LLMs to generate instructions for unlabeled text and rewrite the response, which is essentially simulating how humans would annotate instruction data given candidate unlabeled text.

\section{Method}

The goal of our method is to automatically build instruction data from an unlabeled text corpus.
In this section, we first outline how we select unlabeled texts that contain well-structured helpful and insightful contents (\S \ref{sec:sample_unlabeled_text}). Next, we describe how to convert these contents into <instruction, response> pairs through instruction generation (\S \ref{sec:build_instruction_data}) and response rewriting (\S \ref{sec:response_rewriting}). Finally, we explain how to instruction finetune pretrained LLMs using both seed instruction data and our synthetic data (\S\ref{sec:intstruction_tuning}). Figure \ref{fig:method} illustrates the overall framework of our method.

\subsection{Unlabeled Text Selection}
\label{sec:sample_unlabeled_text}

The first step of our method is to select a subset of unlabeled text rich in high-quality instruction following information to steer the direction of model generation.
Therefore, we aim to find a subset of text that contains well-structured helpful and insightful content to serve as the source of instruction tuning signals of our annotated data. 

While there might be more sophisticated ways to select the desired text, we simply use a set of heuristic rules to filter out low quality text. Using rule-based filtering has its merits compared with advanced neural methods, as it is fast, stable, and interpretable. The exact sampling rules can be found in Appendix \ref{appendix:exact_rules}. Some of the major factors and rules we consider include: 
\begin{enumerate}
    \item \textbf{Text length}, which is generally related to the quality of the text. Serious and complicated topics often require longer text to elucidate, and longer text usually contains more details. Thus we filter out texts that are too short. 
    \item \textbf{Text structure}, which usually indicates the overall quality of the text and the expertise of the author. Well organized content shows the seriousness of writing and professionalism of the author. We keep text that contains multiple paragraphs starting with verbs in imperative form or present participles. 
    \item \textbf{Text style}, as formal writing usually results in high-quality content. We filter out text with too many first-person pronouns such as "I", "my", "we", and "us", as these pronouns are more common in casual writing. 
    \item \textbf{Promotional text indicators}, since text corpus includes lots of advertisement, we filter out text containing punctuation commonly found in advertisement. We also filter out text with too many all-capitalized words, as they are also common in promotional text.
\end{enumerate}

The resulting subset is a diverse set of high-quality unlabeled text, more suitable for converting into instruction response than randomly sampled text. We find this subset mainly consists of helpful guidance for a particular task, provide solutions for certain problem, or elucidate some complicated concepts. Such texts are more likely to have a proper instruction, and help reduce the noise during instruction generation. The natural diversity inherited from unlabeled corpus also resembles our daily needs, which is more suitable for instruction tuning an LLM-based AI assistant. Table \ref{tab:sample} shows an instance of our selected unlabeled text. More examples can be found in Appendix \ref{sec:examples}.

\begin{table}[t]
  \centering
  \resizebox{\columnwidth}{!}{
  \begin{tabular}{p{9cm}} 
  \toprule
    \textbf{Unlabeled Text:}

    Traveling is often a joyous experience, where the only moans and groans come from not being able to see everything you wanted to...
    
    These safety tips will help you travel safely on your final trips of the summer.
    
    Dress moderately. Now is not the time to look like you have money, even if that is the case. Dress moderately, and hide anything flashy...
    
    Travel light. For one, it’s easier to keep with you at all times—deterring thieves by making it less likely that you’ll need to set it down...
    
    ...
    
    When you practice these simple safety tips, you drastically improve the safety of your travels. Remember these tips every time you travel and always be on the lookout for new safety tips. Your family and identity are important, so guard them carefully. \\
    \midrule

    \textbf{Generated Instruction:}

    Give me some safety tips for my final trips of the summer. \\
    \midrule

    \textbf{Rewritten Response:}

    Here are some safety tips for your final trips of the summer:
    
    1. Dress moderately: Avoid drawing attention to yourself by dressing like everyone else. Avoid excessive jewelry or flashy clothing styles that may attract unwanted attention.
    
    2. Travel light: Carrying less weight will make it easier to keep track of your belongings and reduce the chances of becoming a target for thieves.
    
    ...
    
    Remember, these safety tips are meant to enhance your overall safety and protect your family and identity. Stay alert and continue to explore new safety tips to ensure a safe and enjoyable travel experience. \\ \bottomrule
  \end{tabular}
  }
  \caption{An instance of our selected unlabeled text, along with generated instruction and rewritten response. Some texts are omitted for length issues.}
  \label{tab:sample}
\end{table}

\subsection{Instruction Generation}
\label{sec:build_instruction_data}

To obtain instruction-response pairs, we reversely train an LLM on a small set of human-annotated seed instruction data $\mathcal{D}_{\text{seed}}=\{(x_{s}, y_{s})\}$, and use the trained model $\mathcal{M}_{\text{reverse}}$ to add instructions to the selected unlabeled text \cite{self-alignment,longform}.
The generated instruction is answerable by the unlabeled text or by part of it.
This is similar to the back-translation setting in machine translation \cite{back-translation}.
The objective for reversely training $\mathcal{M}_{\text{reverse}}$ is:

\begin{equation}
    \mathcal{L}_{\text{reverse}} = \sum_{(x_{s},y_{s})\in\mathcal{D}_{\text{seed}}}-\log p(x_{s}|y_{s})
\end{equation}
where $x_{s}$ represents instruction and $y_{s}$ represents response in human-annotated seed instruction data.

\subsection{Response Rewriting}
\label{sec:response_rewriting}

Using selected unlabeled text as responses for newly generated instructions is suboptimal. The distribution mismatch between unlabeled text and instruction response results in some noise in the instruction-response pairs, mostly in the form of irrelevant information that does not relate to the newly generated instruction. In addition, the unlabeled text is written from a human perspective, and may differ in the style of a helpful AI assistant. To reduce such noise, we employ a response rewriting process to improve the quality of unlabeled text as instruction response.

\begin{figure}[t]
    \centering
  \resizebox{0.8\columnwidth}{!}{
  \begin{tcolorbox}[
      width=\columnwidth,
      colback=white, 
      colframe=gray!50!black, 
      coltitle=black, 
      fonttitle=\bfseries\large, 
      arc=4mm, 
      enhanced, 
      attach boxed title to top left={yshift=-\tcboxedtitleheight/2, xshift=10pt}, 
      boxed title style={
          enhanced,
          colback=white,
          colframe=white,
          arc=0mm,
          left=0pt,
          right=0pt,
          boxsep=0pt
      }
  ]
  
  {
    Answer the question based on the web text provided. Your answer must be helpful, detailed, and polite. Hide the fact that your answer is actually based on the web text, i.e. answer directly. \\
    
    Web Text: \\
    \texttt{[Selected unlabeled text]} \\
    
    Question: \\
    \texttt{[Reversely generated instruction]} \\
  }
  
  \end{tcolorbox}
  }
  \caption{Prompt for collecting rewritten responses from ChatGPT.}
  \label{fig:rewrite_prompt}
\end{figure}

Our rewriting process is formed similar to generative machine reading comprehension, in which the rewriting model $\mathcal{M}_{\text{rewrite}}$ is given the unlabeled text as the context, the instruction as the question, and is required to generate a response as the answer. To obtain $\mathcal{M}_{\text{rewrite}}$, we train another LLM on a small set of rewriting data $\mathcal{D}_{\text{rewrite}} = \{(x_{g},y_{u},y_{r})\}$, which contains tuples of unlabeled text $y_{u}$, the generated instruction $x_{g}$, and the rewritten response $y_{r}$, i.e. $\mathcal{M}_{\text{rewrite}}$ is trained with the following objective:

\begin{equation}
    \mathcal{L}_{\text{rewrite}} = \sum_{(x_{g},y_{u},y_{r})\in\mathcal{D}_{\text{rewrite}}}-\log p(y_{r}| y_{u}, x_{g})
\end{equation}

To collect $\mathcal{D}_{\text{rewrite}}$ for training rewriting model, we use ChatGPT as a proxy for human annotators to rewrite selected unlabeled texts based on generated instructions. We ask ChatGPT to hide the fact that the response is generated based on some given unlabeled text (see prompt in Figure \ref{fig:rewrite_prompt}).  We find ChatGPT returns valid responses grounded in the given text in most cases, demonstrating the quality of our selected text and the validity of generated instructions. However, there are two cases when ChatGPT fails to rewrite the response properly:

\begin{enumerate}
    \item ChatGPT fails to hide the fact that the response is generated based on the given unlabeled text, and produces responses including text segments such as "Based on the given web text", which is a leakage from our rewrite prompt. These failure cases are filtered out when training the rewriting model.
    \item ChatGPT decides that the unlabeled text does not contain enough information to generate a proper response and refuses to answer the instruction. These responses usually start with "I am sorry" or "I apologize". We keep these failure cases when training rewriting model, as they provide additional defense against noise in our unlabeled text selection and instruction generation. This is similar to the inclusion of unanswerable questions in machine reading comprehension \cite{squad2}. After response rewriting, these cases can be filtered out using simple rules (Appendix \ref{appendix:rewritten_filter_rules}).
\end{enumerate}

Both the instruction generation model and the rewriting model are trained at a small scale, with data size similar to seed instruction data. They are kept fixed when scaling up the size of generated data without further annotation cost. We collect the generated instruction $x_{g}$ and the rewritten response $y_{r}$ to construct our final synthetic data $\mathcal{D}_{synthetic} = \{(x_{g}, y_{r})\}$.
Table \ref{tab:sample} shows an instance of generated instruction and the rewritten response.

\subsection{Joint Fine-tuning}
\label{sec:intstruction_tuning}

For the final instruction tuning, we train LLMs jointly on the seed instruction data and our synthetic data. To balance two sources of training data, we up-sample seed data according to the size of synthetic data, as it proves useful in back-translation \cite{back-translation-at-scale}. We append different prompts to these two types of instruction data for differentiation, similar to the setting of tagged back-translation \cite{tagged-back-translation}. Specifically, we use the same prompt from Humpback \cite{self-alignment}, i.e. we append Prompt $\text{Pr}_{seed}=$ "Answer in the style of AI Assistant." after the seed instructions and append Prompt $\text{Pr}_{aug}=$ "Answer with knowledge from web." after our generated instructions. During inference, we prepend $\text{Pr}_{seed}$ before $\text{Pr}_{aug}$, as is shown to be beneficial for the final performance \cite{self-alignment}. The objective of joint fine-tuning is:

\begin{equation}
\begin{aligned}
    \mathcal{L}_{\text{finetune}} =  
    & \lambda \sum_{(x_{s},y_{s})\in\mathcal{D}_{\text{seed}}}-\log p(y_{s}|\text{Pr}_{seed}, x_{s}) \\
    & +
    \sum_{(x_{g},y_{r})\in\mathcal{D}_{\text{synthetic}}}-\log p(y_{r}|\text{Pr}_{aug}, x_{g})
\end{aligned}
\end{equation}
where $\lambda$ represents upsample ratio of seed data.

\section{Experiments}

\begin{table*}[ht]
  \centering
  \resizebox{0.82\textwidth}{!}{
    \begin{tabular}{@{}llllc@{}}
      \toprule
                             & \textbf{Model}         & \textbf{\#Annotated Data} & \textbf{\#Fine-tuning Data} & \textbf{AlpacaEval(\%)}  \\
      \midrule
      \textbf{Proprietary}   & GPT-3.5-Turbo-0301     & -                       & -                 & 89.37                    \\
                             & GPT-4                  & -                       & -                 & 95.28                    \\
      \midrule
      \textbf{Distilled}     & Baize-v2-7B            & 100k                    & 100k              & 63.85                    \\
                             & Vicuna-7B-v1.1         & 70k                     & 70k               & 64.41                    \\
                             & Vicuna-7B-v1.3         & 125k                    & 125k              & 76.84                    \\
      \midrule
      \textbf{Non-distilled} & Guanaco-7B             & 9k                      & 9k                & 46.58                    \\
                             & Dromedary$_{ours}$     & 195                     & 40k               & 18.52                    \\
                             & Humpback$_{ours}$      & 3k                      & 3k+32k            & 49.19                    \\
                             & OASST-7B$_{3K}$        & 3k                      & 3k                & 61.70                    \\
                             \midrule
     \textbf{Ours}           & REInstruct-7B$_{8K}$  & 3k+4k                   & 3k+8k            & 63.78           \\
                             & REInstruct-7B$_{16K}$  & 3k+4k                   & 3k+16k            & 64.75           \\
                             & REInstruct-7B$_{32K}$  & 3k+4k                   & 3k+32k            & \textbf{65.41}           \\
      \bottomrule
    \end{tabular}
  }
  \caption{Win rate of instruction tuned LLMs against text-davinci-003 on AlpacaEval leaderboard.}
  \label{tab:sft_results}
\end{table*}

\begin{table*}[ht]
  \centering
  \resizebox{0.66\textwidth}{!}{
    \begin{tabular}{@{}lccccc@{}}
      \toprule
      \textbf{Model}           & \textbf{Hellaswag}  & \textbf{PIQA}     & \textbf{Winogrande}   & \textbf{ARC\_E}   & \textbf{ARC\_C}   \\
      \midrule
      Baize-v2-7B              & 73.21               & 79.54             & 68.98                 & 69.44             & 44.37             \\
      Vicuna-7B-v1.1           & 74.64               & 78.62             & \textbf{70.17}        & 72.01             & 43.77             \\
      Vicuna-7B-v1.3           & 73.92               & 79.22             & 69.30                 & 71.89             & 44.45             \\
      REInstruct-7B$_{32K}$    & \textbf{77.04}      & \textbf{79.76}    & 69.77                 & \textbf{73.19}    & \textbf{48.04}    \\
      \bottomrule
    \end{tabular}
  }
  \caption{Results (Accuracy) of different instruction-tuned models on commonsense reasoning.}
  \label{tab:qa_results}
\end{table*}


\subsection{Setup}

\textbf{Dataset.} We use C4 \cite{t5} as our unlabeled text corpus, since it is a free open-source dataset commonly used in pre-training LLMs. C4 contains around 400M text segments cleaned from Common Crawl Project\footnote{https://commoncrawl.org}. For the seed instruction data used in training instruction generation model and final instruction tuning, we use 3k high-quality (rank 0) instruction data obtained from the first round of conversations in OpenAssistant \cite{oasst}. We collect 4k rewriting data from gpt-3.5-turbo for training the rewriting model.

\textbf{Baselines.} We mainly compare our method with other instruction tuned models that do not rely on powerful proprietary models:

\begin{itemize}
    \item OASST$_{3K}$ \cite{oasst}: Llama model tuned with our 3k seed instruction data.
    \item Guanaco \cite{guanaco}: Llama model tuned with 9k instruction data from all turns in OpenAssistant dataset.
    \item Dromedary$_{ours}$ \cite{dromedary}: Llama model trained with 40k randomly sampled instruction data from the original Dromedary dataset.
    \item Humpback$_{ours}$ \cite{self-alignment}: We re-implemented Humpback using C4 and Llama-7b under our settings. More details and discussions can be found in Appendix \ref{appendix:humpback}.
\end{itemize}

We also compare with models trained with instruction data distilled from proprietary LLMs \cite{vicuna, wizardlm, baize} to show the competitiveness of our method.

\textbf{Evaluation.} We use AlpacaEval \cite{alpaca_eval} to evaluate the instruction following capability of our instruction tuned LLMs. AlpacaEval consists of 805 test instructions that covers a variety of tasks. The evaluation metric is the win rate against reference response by text-davinci-003, judged by GPT-4 \cite{gpt4}.

\textbf{Implementation Details.}
We use Llama-7b \cite{llama} as our pre-trained base LLM for data construction, which is fine-tuned for instruction generation and response rewriting. For training these two models, we use a batch size of 128, a learning rate of 2e-5 and a cosine learning rate schedule for 10 epochs, which gradually reduces the learning rate to zero. The up-sample \cite{back-translation-at-scale} ratio for seed data during joint-tuning is chosen according to the growing size of newly generated data, keeping the ratio of generated data to seed data roughly 2:1 (see Appendix \ref{appendix:finetune_details}). All instruction tuning experiments are conducted with the same batch size of 128 and a constant learning rate of 2e-5. We use 4 or 8 A100-80G GPUs for most of our training and carry out full-parameter tuning.
During instruction generation and response rewriting, we use greedy decoding with repetition penalty \cite{repetition_penalty_1,repetition_penalty_2} of 1.05.

\begin{figure*}[ht]
    \begin{subfigure}[b]{\columnwidth}
    \centering
    \includegraphics[width=0.92\columnwidth]{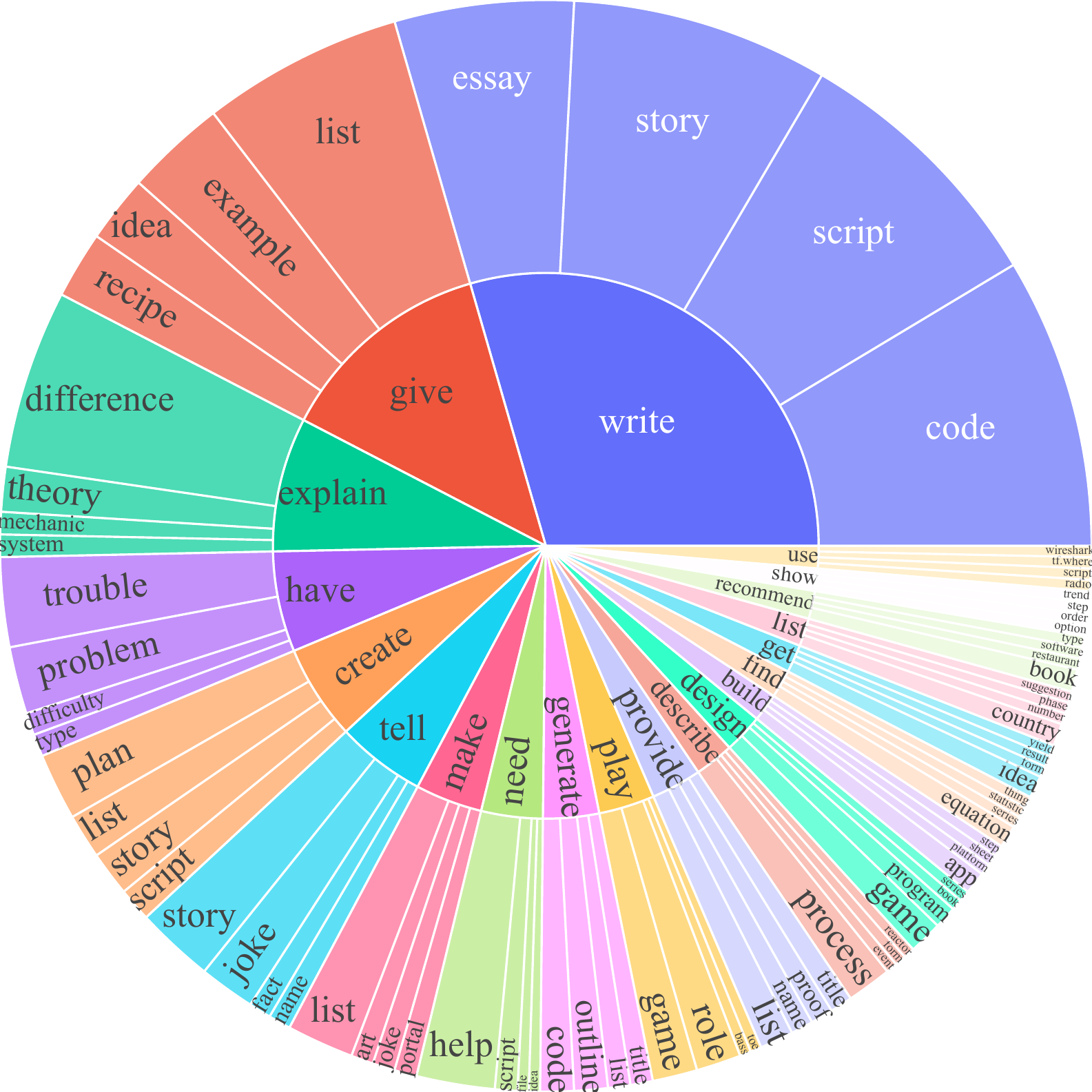}
    \caption{seed instructions}
    \label{fig:seed_sunburst}
    \end{subfigure}%
    ~
    ~
    ~
    ~
    \begin{subfigure}[b]{\columnwidth}
    \centering
    \includegraphics[width=0.92\columnwidth]{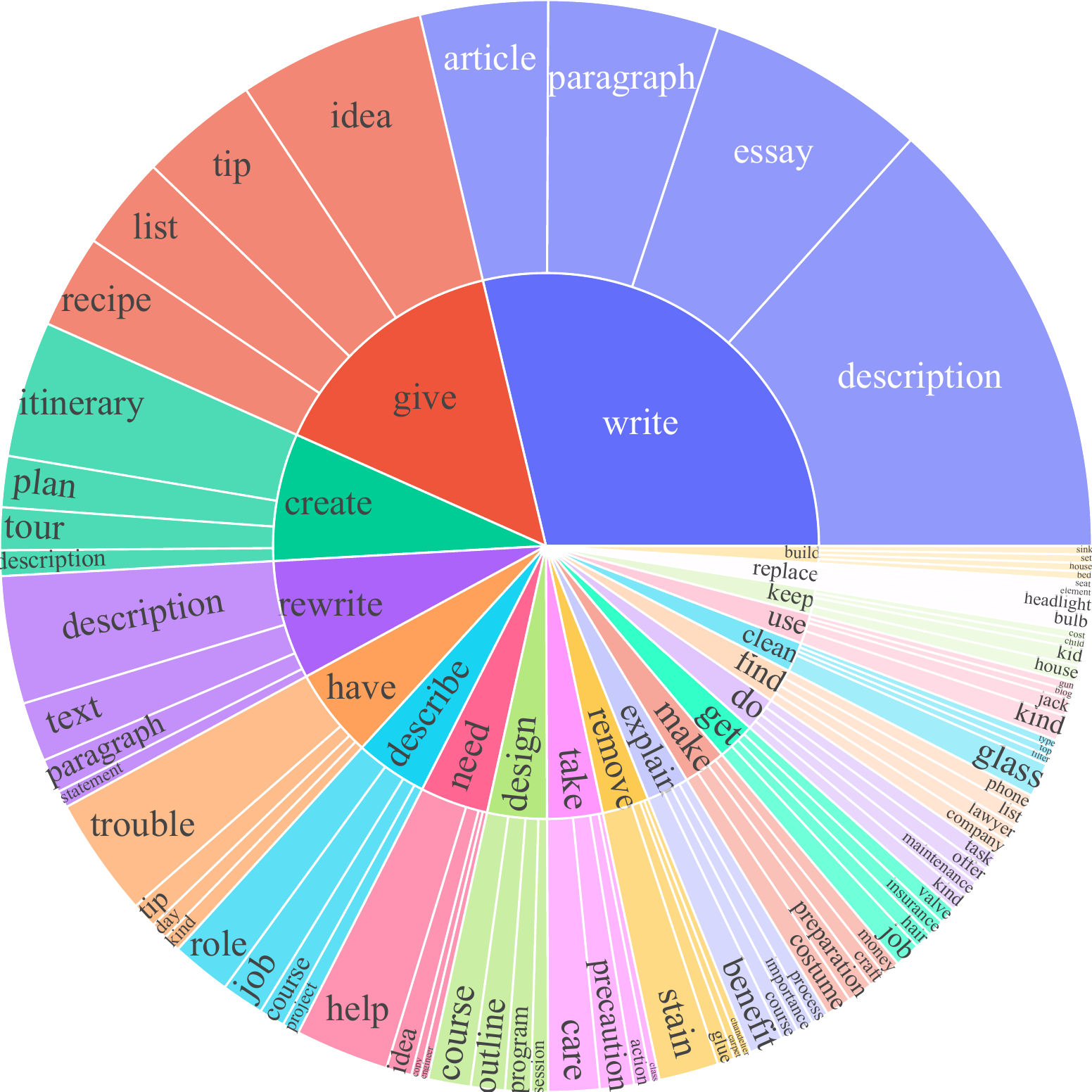}
    \caption{generated instructions}
    \label{fig:generated_sunburst}
    \end{subfigure}%
    \caption{Sunburst visualization of verb-noun structure of seed instructions and generated instructions on our selected unlabeled text.}
    \label{fig:sunburst_comparison}
\end{figure*}

\subsection{Main Results}

We select unlabeled text of size 8k, 16k, and 32k for our experiments, and train Llama-7b jointly on seed data and our synthetic data across different data scales.
Table \ref{tab:sft_results} shows the results of our method and other models on AlpacaEval.
We can see that:

1) \textit{REInstruct can achieve better performance than other open-source, non-distilled methods.}
Our method, which learns from only a few seed data and automatically built instructions, surpasses most non-distilled methods and achieves competitive performance with some distilled methods, such as Baize-v2-7B and Vicuna-7B-v1.1.

2) \textit{By automatically building high-quality instruction data from an unlabeled corpus, REInstruct can lead to scalable improvement on instruction following.}
Experiments with 8k, 16k, and 32k data show continuous performance gains as data volume grows.

To better verify the effectiveness of our model, we further compare its performance with other distilled models on commonsense reasoning.
The evaluation is conducted using lm-eval \cite{eval-harness}.
As shown in Table \ref{tab:qa_results}, our method achieves better or competitive performance on five commonsense reasoning datasets.

\subsection{Quality Analysis of Auto Generated Data}

This section analyzes the quality of automatically generated data, including generated instructions and rewritten responses.

\subsubsection{Diversity of Generated Instruction}
Since our seed instruction data consists of just over 3k data points, we select a subset of 4k unlabeled text from C4 for instruction generation, maintaining a similar data size. 
Figure \ref{fig:sunburst_comparison} shows the sunburst visualization of verb-noun structure of seed instructions and our generated instructions, as used in Self-Instruct \cite{self-instruct} to show the diversity of tasks.
We found our generated instructions are quite diverse and similar to the seed instructions, covering a wide range of queries that may be sent to an AI assistant for help or guidance.

\subsubsection{Similarity of Unlabeled Text and Rewritten Response}

To show that our rewriting process generates responses based on the unlabeled text and preserves the instruction following signals in it, we measure text similarity between the unlabeled text and the rewritten response. Specifically, we calculate the ratio of words in rewritten response that can be directly copied from unlabeled text. We found that 77.63\% of the words in rewritten response also appear in the unlabeled text.

\begin{figure}[t]
  \centering
  \includegraphics[width=0.88\columnwidth]{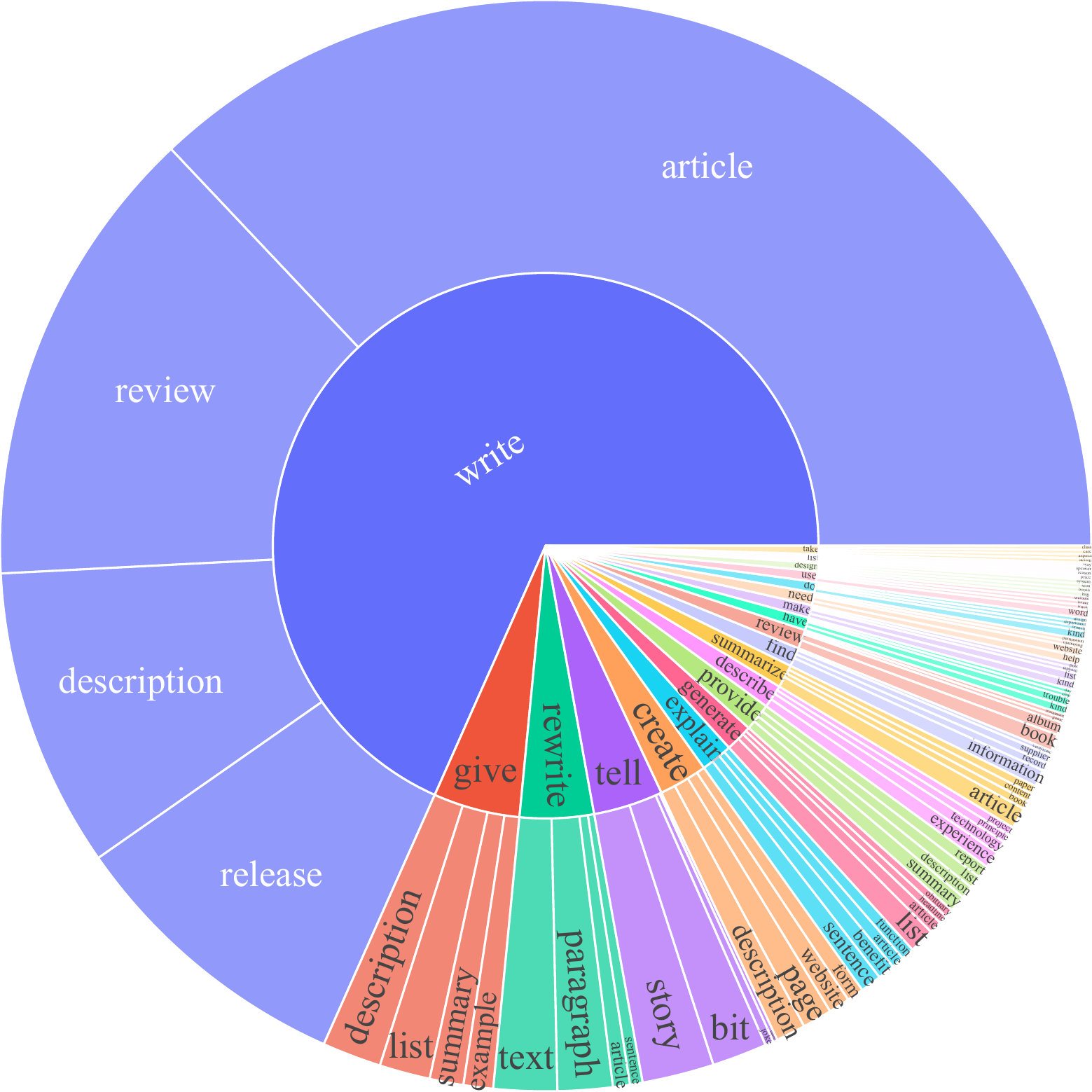}
  \caption{Sunburst visualization of verb-noun structure of generated instructions on randomly sampled unlabeled text.}
  \label{fig:c4_random_4000}
\end{figure}

\subsection{Detailed Analysis}
This section analyzes the effects of Unlabeled Text Selection and Response Rewriting.
We design four ablated variants of REInstruct:
\begin{itemize}
    \item ``w/o Unlabeled Text Selection'' directly samples unlabeled text from the corpus at random, rather than using several heuristic rules to obtain high-quality texts (\S \ref{sec:effect_of_unlabeled_text_selection});
    \item ``w/o Response Rewriting'' removes the response rewriting step and directly uses unlabeled text as the response (\S \ref{sec:effect_of_response_rewriting});
    \item ``Self-Rewrite'' uses the LLM only trained on seed data to rewrite the unlabeled text (\S \ref{sec:effect_of_response_rewriting});
    \item ``Self-Training'' uses the self-generated response instead of the rewritten response from unlabeled text (\S \ref{sec:effect_of_instruction_following_signals});
\end{itemize}

We conduct analysis at the scale of 8k synthetic data, and Table \ref{tab:ablation} shows the win rate on AlpacaEval of these different variants.

\begin{table}[t]
  \centering
  \resizebox{0.99\columnwidth}{!}{
    \begin{tabular}{@{}lc@{}}
      \toprule
                         & \textbf{AlpacaEval(\%)} \\
      \midrule        
      OASST-7B$_{3K}$   & 61.70                   \\
      \midrule        
      REInstruct-7B$_{8K}$    & 63.78                   \\
      \ \ \ w/o Unlabeled Text Selection & 61.42                   \\ 
      \ \ \ w/o Response Rewriting   & 49.19                   \\
      \midrule        
      Self-Rewrite      & 56.63                   \\
      Self-Training     & 59.76                   \\
      \bottomrule
    \end{tabular}
  }
  \caption{\label{tab:ablation}Win rate on AlpacaEval of REInstruct and its ablated variants.}
\end{table}

\subsubsection{Effect of Unlabeled Text Selection}
\label{sec:effect_of_unlabeled_text_selection}

To understand the effect of our unlabeled text selection, we compare it with random sampling from the text corpus. Figure \ref{fig:c4_random_4000} shows the sunburst visualization of instructions generated on randomly sampled unlabeled text. As seen, most instructions are limited to a few text generation tasks, indicating low instruction diversity.

We then conduct an instruction tuning experiment and evaluate on AlpacaEval, using the same process as the original REInstruct but with the randomly sampled unlabeled text.
Table \ref{tab:ablation} shows that using randomly sampled unlabeled text (w/o Unlabeled Text Selection) leads to no significant performance improvement over the baseline OASST-7B$_{3K}$ (61.70\% to 61.42\%), while our method REInstruct-7B$_{8K}$ improves performance from 61.70\% to 63.78\%.

Based on the above observation, it is clear that not all unlabeled texts are worth converting into instruction tuning data. And our unlabeled text selection can effectively obtain high-quality unlabeled text for constructing instruction-following data.

\subsubsection{Effect of Response Rewriting}
\label{sec:effect_of_response_rewriting}

To show the necessity of the rewriting process, we conduct instruction tuning without rewriting unlabeled text, i.e. the selected unlabeled text is directly used as instruction response.
As shown in Table \ref{tab:ablation}, using unlabeled text as responses (w/o Response Rewriting) significantly degrades performance.
This observation indicates that the content from unlabeled texts are unsuitable to be directly used as accurate instruction responses for model training.

An interesting question is whether we can skip with the external rewriting data and directly employ the LLM instruction-tuned on the seed data to rewrite the unlabeled text.
We refer such approach as Self-Rewrite.
Table \ref{tab:ablation} shows that self-rewrite leads to inferior performance.
We suspect that LLMs fine-tuned on a small set of seed data with relatively short instructions might struggle with the rewriting task where the rewriting instructions are much longer. Adding longer instructions or other content-grounded generation data to the seed data might help reduce reliance on external rewriting data. We will explore this in future work.

\subsubsection{Effect of Instruction Following Signals in Unlabeled Text}
\label{sec:effect_of_instruction_following_signals}

In this section, we investigate the role of instruction following signals extracted in our selected unlabeled text. Previous works \cite{lima} have shown that LLMs obtain quite reasonable instruction following capability with limited instruction data. Therefore, it is possible that a self-training approach where model is trained on its own responses on a larger set of instructions may lead to better performance.

To understand the difference between self-training and our method, we use instructions reversely generated on our unlabeled text to prompt LLM tuned on seed data. We then finetune a new LLM jointly on the seed data and the self-generated data. In other words, the sampled unlabeled text is only used to produce a diverse set of instructions, and does not provide instruction following signals in the response.

Table \ref{tab:ablation} shows that the result of self-training is inferior to our method. Our hypothesis is that good instruction tuning requires continuously providing high quality instruction following signals that can improve current output space of the trained LLM. The instruction data generated in a self-training fashion already holds large amount of probability space, and does not provide much training signals.

\section{Conclusion}

In this paper, we propose REInstruct, a simple and scalable method to automatically build instruction data from unlabeled corpus. Our method only requires small amount of labeled data, and does not heavily rely on proprietary LLMs. Experiment results show that instruction tuning LLMs on a small set of human-labeled data and our synthetic data achieves competitive instruction following performance, demonstrating the quality of our synthetic data. We believe this work provides insights for instruction data annotation and helps foster research in instruction tuning. For future work, we aim to develop more sophisticated neural methods to collect and extract instruction following signals from unlabeled dataset across different modalities.

\section*{Limitations}

Using heuristic rules to sample unlabeled text has certain limitations. First, it only captures some shallow textual features that are indicators of high-quality text, with limited semantic understanding of the content. Second, it might be hard to generalize to other data distributions such as code, or other modalities such as vision. Finding high-quality unlabeled data might be more subtle in these settings.

While our method leads to more capable LLMs, further alignment training is required to reduce the hallucination and toxicity of the instruction tuned LLMs.

\section*{Acknowledgements}

We sincerely thank all anonymous reviewers for their insightful comments and valuable suggestions. This research work is supported by the National Natural Science Foundation of China under Grants no. 62122077, no. 62306303, no. 62106251, and no. 62076233.

\bibliography{custom}

\clearpage
\appendix
\onecolumn

\section{Implementation Details}

\subsection{Heuristic Rules to Sample Unlabeled Text}
\label{appendix:exact_rules}

We describe the factors we consider when devising the heuristic rules to select candidate unlabeled texts in section \ref{sec:sample_unlabeled_text}. Here are the exact rules:

\begin{enumerate}
  \item Texts shorter than 1200 characters or longer than 3000 characters are filtered out.
  \item Texts will be kept if there are 4 to 10 paragraphs starting with verbs in imperative form or present participles, and the number of other paragraphs is less than 2.
  \item Texts will be filtered out if they contain more than two words from this list: "we ", "our ", "i ", "i've ", "we've ", "we're ", "my ", "he ", "she ", "us ".
  \item Texts with the following punctuation will be filtered out: "...", "…", "™", "\#", "\&", "*", "®", "@".
  \item Texts with more than two all-capitalized words such as "COME" and "NOW" are filtered out.
  \item Texts with more than one question mark("?") will be filtered out.
\end{enumerate}

\subsection{Rules to Filter Failed Rewritten Data}
\label{appendix:rewritten_filter_rules}

To filter out responses that fail to hide the fact that the answer is based on the given web text, we remove responses that contain "web text"(leakage from our prompt) and "based on the information provided". To filter out responses that refuse to answer the instruction, we remove responses that contain "sorry" and "i apologize".

\subsection{Finetune Details}
\label{appendix:finetune_details}

For instruction tuning on seed data, we use batch size of 128 and train Llama-7b for 200 steps. For joint instruction tuning on seed data and synthetic data, we also use batch size of 128 and finetune models with the number of steps and upsample ratio listed in Table \ref{tab:train_steps_and_upsample_ratio}.

\begin{table}[ht]
  \centering
    \resizebox{0.8\columnwidth}{!}{
      \begin{tabular}{@{}lllcc@{}}
        \toprule
        \textbf{Model}        & \textbf{\#Annotated Data} & \textbf{\#Fine-tuning Data} & \textbf{Training Steps} & \textbf{Upsample Ratio} \\
        \midrule
        REInstruct-7B$_{8K}$  & 3k+4k                   & 3k+8k             & 500      & 1            \\
        REInstruct-7B$_{16K}$ & 3k+4k                   & 3k+16k            & 800      & 2            \\
        REInstruct-7B$_{32K}$ & 3k+4k                   & 3k+32k            & 1000     & 4            \\
        \bottomrule
      \end{tabular}
    }
  \caption{\label{tab:train_steps_and_upsample_ratio}Training steps and upsample ratio}
\end{table}

\subsection{Prompts}
\label{appendix:exact_prompt}

\begin{figure}[!th]
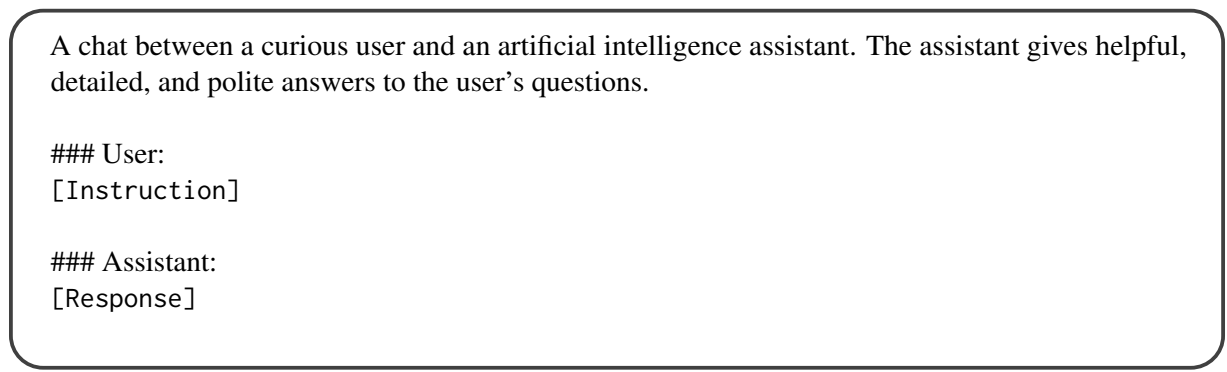

  \begin{tcolorbox}[
      width=\columnwidth,
      colback=white, 
      colframe=gray!50!black, 
      coltitle=black, 
      fonttitle=\bfseries\large, 
      arc=4mm, 
      enhanced, 
      attach boxed title to top left={yshift=-\tcboxedtitleheight/2, xshift=10pt}, 
      boxed title style={
          enhanced,
          colback=white,
          colframe=white,
          arc=0mm,
          left=0pt,
          right=0pt,
          boxsep=0pt
      }
  ]
  
  {
    A chat between a curious user and an artificial intelligence assistant. The assistant gives helpful, detailed, and polite answers to the user's questions. \\
    
    \#\#\# User: \\
    \texttt{[Instruction]} \\
    
    \#\#\# Assistant: \\
    \texttt{[Response]} \\
  }
  
  \end{tcolorbox}
  \caption{System prompt for instruction tuning}
  \label{fig:sft_prompt}
\end{figure}

Figure \ref{fig:sft_prompt} shows the system prompt we use for instruction tuning.

\clearpage
\section{Our implementation of Humpback}
\label{appendix:humpback}

We use the exact scoring prompt in Humpback \cite{self-alignment}, and run the iterative curation process on our 32k selected unlabeled text. The base model we use is Llama-7b. We found that during the first iteration, about half of our unlabeled text is ranked 5, and the other half is ranked 4. We use texts with score 5 to train the model in the second iteration, and the resulting model produces score 5 for almost all our unlabeled text, thus the performance of the final trained model will be the same as our ablation experiment "w/o Response Rewriting" in Table \ref{tab:ablation}, which has win rate about 50\%, much lower than the performance reported in the Humpback paper. Possible reasons for such performance are:

\begin{enumerate}
    \item Difference in implementation: Since Humpback is not open-source, some exact details are unclear. For example, it is possible that they use Llama-65b model to curate instruction data and use the obtained data to train Llama-7b model for ablation. We also notice other self-alignment works \cite{dromedary,self-rewarding} typically use larger LLMs of scale 65b or 70b.
    \item Difference in unlabeled dataset: Since ClueWeb \cite{clueweb} used in Humpback \cite{self-alignment} is not a free open-source dataset, and we are unable to obtain it in a short time, we use the more accessible C4 dataset for our experiment. The difference between CommonCrawl and ClueWeb is well-discussed in \citet{clueweb22}, and ClueWeb generally contains higher quality texts that are more likely to satisfy potential information needs.
\end{enumerate}

To the best of our knowledge, there are no public implementations of Humpback on the 7b model scale. On the other hand, it is completely possible to combine self-alignment works such as Humpback \cite{self-alignment} with our method. We leave that for future investigation.

\section{Possible Bias from GPT-4 Evaluation}

In the case of AlpacaEval, three potential biases of automatic evaluators such as GPT-4 are:

\begin{itemize}
    \item Prefer longer responses: It is possible that longer responses contribute to the win rate of our method, but since our average response length is not unreasonably longer than other models (Table \ref{tab:length_bias}), we believe our results are still valid in the comparison setting.
    \item Gives more importance to the style of the response than its content (e.g. factuality): More investigation might be needed to understand how this bias affects our reported results. It is possible that our responses are in a more helpful or friendly tone (e.g. trying to provide more information), but has some issues such as factuality.
    \item Prefer responses from models that are similar: Since we do not directly distill responses from ChatGPT or GPT-4, this bias has less impact on our reported win rate.
\end{itemize}

\begin{table*}[ht]
  \centering
  \resizebox{0.64\columnwidth}{!}{
    \begin{tabular}{@{}lllc@{}}
      \toprule
                             & \textbf{Model}             & \textbf{Average Length}  & \textbf{AlpacaEval(\%)}  \\
      \midrule
      \textbf{Distilled}     & Baize-v2-7B                & 1127                     & 63.85                    \\
                             & Vicuna-7B-v1.1             & 1127                     & 64.41                    \\
                             & Vicuna-7B-v1.3             & 1100                     & 76.84                    \\
      \midrule
      \textbf{Non-distilled} & Guanaco-7B                 & 1364                     & 46.58                    \\
                             & OASST-7B$_{3K}$            & 991                      & 61.70                    \\
                             & REInstruct-7B$_{32K}$      & 1096                     & \textbf{65.41}           \\
      \bottomrule
    \end{tabular}
  }
  \caption{Average response length on AlpacaEval test data.}
  \label{tab:length_bias}
\end{table*}

\section{Examples}
\label{sec:examples}

We found almost every unlabeled text we selected contains some useful information or knowledge. However, it is possible that average people like us are not capable to judge whether the text contain the best information or knowledge about the discussed topic. Table \ref{tab:negative_sample_1} provides an example that we think might not be good enough. Meanwhile, some texts that are suitable for converting into instruction data might be filtered out by our heuristic rules. We consider finding such texts a major future direction. One possible solution is to use the selected texts as positive examples and random texts as negative examples to train a neural classifier. Table \ref{tab:negative_sample_2} shows an example that might be good for converting into instruction data.

We include more <unlabeled text, generated instruction, rewritten response> triples in Table \ref{tab:example_1} and Table \ref{tab:example_2}, so that readers can better understand our synthetic data.

\clearpage
\begin{table*}[!ht]
  \centering
  \resizebox{0.8\columnwidth}{!}{
  \begin{tabular}{p{14cm}} 
  \toprule
  Being An Entrepreneur is not at all Glamorous It’s stressful. If you think meeting your boss’s deadlines or demands is tough, try meeting your own, especially when your personal savings are on the line, workload can be intense and it’s frustrating.
  
  Remember the time when you had the great idea to start your business? You did everything possible to turn that vision into a reality. It was exciting, nerve-wracking and rewarding all at the same time.
  
  Don’t fear failure; people we try to avoid the negative consequences of failure; BUT we also lose out on the chance of success.
  
  Create a wall of positivity, a motivational wall start printing out quotes, pictures that motivates you and reminds you of your passion.
  
  Surround yourself with like-minded people. Try to spend time with people who have a positive outlook on life, they can be very inspiring, and avoid negative people.
  
  Take care of yourself, recharge your energy by doing something that makes you happy.
  
  Have other goals to achieve in your life other than work, that will make you happy and gives you strength and confidence.
  
  Remember you have adopted a lifestyle of taking chances, while someone else has declined that challenge, don’t lose your entrepreneurial spirit. It’s what got you where you are in the first place, and it will help you every day as you live in this non-stop world. \\  \bottomrule
  \end{tabular}}
  \caption{An example of selected unlabeled text that might not be suitable for converting into instruction data}
  \label{tab:negative_sample_1}
\end{table*}

\begin{table*}[!ht]
  \centering
  \resizebox{0.8\columnwidth}{!}{
  \begin{tabular}{p{14cm}} 
  \toprule
  When considering the legacy you leave behind, have you thought of how your leadership skills will affect that legacy? Leadership skills are a core component of a long lasting legacy. Here are some tips for improving your leadership skills.
  
  Leadership begins and ends with vision. People enjoy following someone that has vision and is able to communicate that vision in easy-to-understand ways. This may be done in story telling or in personal interactions with your team. Each leader has their own personality and leadership style, so use your strengths to communicate your vision.
  
  It’s easy to spot a person that is passionate about something. It comes across in the way they talk about it, the level of detail in which they explain it and through their facial and body expressions. Leaders share their passion with others and others respond by aligning their activities to support that passion driven vision. If you’re excited about something, others will be too.
  
  It’s important to practice what you preach. If you don’t model the morals and integrity you are asking of others, it will undermine your leadership. If you’re going to talk the talk, you should walk the walk. You will quickly find that others will begin to emulate your behavior so be sure that you are behaving in a way that you want your team to behave.
  
  You should surround yourself with people that are better than you. Use their strengths to your advantage, challenge them to be even better, and reward them as they excel.
  
  If you expect people to excel, you need to communicate progress along the way. Communicate daily with your inner circle to ensure that their progress is properly aligned with your vision. Be transparent, talk openly about things that are working and things that aren’t. Analyze everything to discover things that can be improved. Pivot when things are not working well. Communicate with team members in your outer circle on a monthly basis. Let them know how things are progressing, talk about things that are working and things that aren’t. A well informed team will always perform better than those that are held in the dark. \\  \bottomrule
  \end{tabular}}
  \caption{An example of filtered out unlabeled text that might be suitable for converting into instruction data}
  \label{tab:negative_sample_2}
\end{table*}

\begin{table*}[]
  \centering
  \resizebox{\columnwidth}{!}{
  \begin{tabular}{l|p{14cm}} 
  \toprule
    \textbf{Unlabeled Text}       & Scratched windowpanes can diminish the beauty of the windows in your home. Learn how to remove these marks with the following tips. Once the glass is damage free and clean, carefully maintain the windowpanes so that they do not become scratched in the future.
    
    Apply a small amount of soapy water to a damp sponge. Wipe each window pane with the sponge to remove dirty residue. The residue may be covering some of the scratches, so removing it will make it easier for you to spot all of the damage on each windowpane. Rinse the sponge out with plain water and move it over the same spots. Put on a pair of rubber gloves and a face mask to protect your skin from chemicals and prevent breathing in the ammonia.
    
    Add equal amounts of ammonia and warm water to a spray bottle. Shake the contents. Spray the surface of each windowpane with the cleaning solution. Wipe each windowpane off with a lint-free cloth. The ammonia will eliminate any fine particles that remain on the glass panes. Dry each pane off with another lint-free cloth.
    
    Pour a few drops of cerium oxide onto a buffing pad. Cerium oxide is a type of polish that is often used to clean jewelry. Place the pad directly over one of the scratches. Press down on the back of the pad as you move it around with small, circular motions. Add a few more drops of cerium oxide as needed. The scratch will fade as you polish it.
    
    Remove the remaining scratches in the same manner. Spray the window panes, once again, with the ammonia and water solution. Eliminate residue that the cerium oxide left behind with a lint-free cloth. Dry each pane well with another cloth. The windowpanes will be damage-free.
    
    Protect the glass from becoming scratched in the future by always using soft cloths and nonabrasive cleaning supplies. Remove dirty residue from the glass as soon as it becomes apparent to prevent it from drying on the glass and damaging its surface. When you open or close a window in your home, grasp the frame that surrounds the glass rather than pushing up or down on the pane. These simple steps will protect the windowpanes so that they continue to look great for years to come.
    
    If you have cracks or holes in your windows, contact a company like Action Glass to repair them for you.  \\ \midrule
    \textbf{Generated Instruction} & How do I remove scratches from windowpanes? \\ \midrule
    \textbf{Rewritten Response}      & To remove scratches from windowpanes, you can follow these steps:
    
    1. Start by applying a small amount of soapy water to a damp sponge. Wipe each window pane with the sponge to remove dirty residue. This will help reveal any scratches that may be hidden underneath.
    
    2. Rinse the sponge out with plain water and repeat the process on the same spots. It's important to wear rubber gloves and a face mask to protect yourself from chemicals and prevent breathing in ammonia.
    
    3. Create a cleaning solution by mixing equal amounts of ammonia and warm water in a spray bottle. Shake the contents and spray the surface of each windowpane with the cleaning solution. Use a lint-free cloth to wipe each windowpane off.
    
    4. To further remove any remaining particles, add a few drops of cerium oxide onto a buffing pad. Cerium oxide is a type of polish commonly used to clean jewelry. Place the pad directly over a scratch and press down on it while moving in small, circular motions. Repeat this step if needed and add more drops of cerium oxide as required.
    
    5. Finally, remove any remaining residue left by the cerium oxide with a lint-free cloth. Make sure to dry each pane thoroughly with another lint-free cloth.
    
    By following these steps, you should be able to effectively remove scratches from your windowpanes and restore their beauty. Additionally, it's important to protect the glass from becoming scratched in the future by using soft cloths and nonabrasive cleaning supplies. Always remove dirty residue as soon as possible to prevent it from damaging the surface of the glass. When opening or closing a window, grasp the frame instead of pushing on the pane, as this can help protect the windowpanes. \\ \bottomrule
  \end{tabular}}
  \caption{An example of selected unlabeled text, along with generated instruction and rewritten response}
  \label{tab:example_1}
\end{table*}

\begin{table*}[]
  \centering
  \resizebox{\columnwidth}{!}{
  \begin{tabular}{l|p{14cm}} 
  \toprule
    \textbf{Unlabeled Text}       & If you are going under the knife then you will need to prepare yourself for any cosmetic surgery procedure in Oregon. Preparing your home, your life, and yourself for plastic surgery will help to make the entire process less stressful. Here are some tips that you should consider.
    
    Talk to your doctor about recommendations for the specific procedure that you have chosen. For example, anyone who has chosen to have a tummy tuck may need to lose a certain amount of weight before surgery. Those who are having breast implants may need to stop taking hormonal drugs. Depending upon the specific surgery you have chosen you may need to prepare in certain ways.
    
    Stop drinking alcohol. Talk with your doctor regarding his policy on smoking cigarettes. By doing this you will clear your system of nicotine and this will help to facilitate the healing process. It is also important that you avoid taking aspirin and any other types of restricted drugs at least two weeks before you have surgery.
    
    Prepare your home properly. You may need to rest up before while afterwards and therefore you need to prepare yourself for this. Make sure you fill prescriptions in advance, make sure that you have clean linens, find somebody to walk your dog and take care of your pets, and so on.
    
    Stock up your fridge as well. When you get back from the hospital the last thing that you really want to have to worry about is going out and getting food. Instead, that all necessary food up in advance and get loads of healthy snack foods that you can use to help you rehabilitate.
    
    Plan exactly how you intend to get to the hospital and back again once the surgery has been done. While you could use a taxi, it would be better to get a friend or family member to make sure you get there and back again safely. It is better to have someone drive you back home and stay with you the night of surgery.
    
    Next Post Next Can Phoenix Dentists Show Good Habits To Promote Healthy Teeth? \\  \midrule
    \textbf{Generated Instruction} & What preparation should I do if I am going under the knife in Oregon? \\ \midrule
    \textbf{Rewritten Response}      & If you are going under the knife in Oregon, there are several preparations you should consider to make the process less stressful. Firstly, it is important to talk to your doctor about any recommendations they may have for your specific procedure. This could include weight loss, stopping certain medications, or any other precautions that may be necessary.
    
    Additionally, it is crucial to stop drinking alcohol and discuss your doctor's policy on smoking cigarettes. This will help clear your system of nicotine and promote a faster healing process. It is also advisable to refrain from taking aspirin and other restricted drugs at least two weeks before the surgery.
    
    Preparing your home is another important aspect of the process. You may need to rest and recuperate both before and after the surgery, so it is essential to have everything in order. This includes filling prescriptions in advance, ensuring you have clean linens, arranging for someone to take care of your pets, and planning for any necessary household tasks.
    
    Furthermore, it is recommended to stock up your fridge with necessary food items before the surgery. Having easy-to-prepare meals and healthy snacks available will save you the hassle of having to leave the house immediately after the procedure.
    
    Lastly, it is crucial to plan your transportation to and from the hospital. While you could rely on public transportation or a taxi, it is safer to have a friend or family member drive you there and back. It is also advisable to have someone stay with you overnight after the surgery for support and assistance.
    
    By following these preparation steps, you can ensure a smoother experience during your cosmetic surgery in Oregon. \\ \bottomrule
  \end{tabular}}
  \caption{An example of selected unlabeled text, along with generated instruction and rewritten response}
  \label{tab:example_2}
\end{table*}

\end{document}